\documentclass{article}

\usepackage[preprint]{neurips_2026}
\workshoptitle{New in Machine Learning (NewInML)}

\usepackage[utf8]{inputenc} 
\usepackage[T1]{fontenc}    
\usepackage{hyperref}       
\usepackage{url}            
\usepackage{booktabs}       
\usepackage{array}          
\usepackage{amsfonts}       
\usepackage{amsmath}
\usepackage{nicefrac}       
\usepackage{microtype}      
\usepackage{xcolor}         
\usepackage{graphicx}
\usepackage{tikz}
\usepackage{needspace}     

\title{When Tools Get in the Way: The Effect of Unnecessary Tool Availability on LLM Answering}

\author{%
  Saanvi Paturi\thanks{Work done while at Spark AI Research.} \\
  UWCSEA East Campus \\
  \And
  Arsen Kenzhebayev\footnotemark[1] \\
  Haileybury Astana \\
  \And
  Arham Sethi\footnotemark[1] \\
  The Shishukunj International School \\
  \AND
  Vyas Raina\thanks{Correspondence: \texttt{vyas@sparkairesearch.com}} \\
  Apta AI \& Spark AI Research \\
  \And
  Ivaxi Sheth \\
  Spark AI Research \\
  \And
  Vatsal Raina \\
  Apta AI \& Spark AI Research \\
}

\begin{document}

\maketitle

\begin{abstract}
Large language models (LLMs) are increasingly deployed with external tools that extend what they can do beyond their own knowledge. Tools help on tasks that need external information, but their availability may also change how a model handles questions that do not need them. Prior work has mostly asked whether models select and use tools appropriately; whether an unnecessary tool changes the correctness of answers has received less attention. We ask whether making a related but unnecessary tool available affects a model's ability to answer from its own knowledge, and whether a preceding tool interaction changes this behaviour. We construct 500 query pairs across 10 knowledge domains. Each pair consists of a tool query, which needs the domain's tool, and a closed-domain query, which does not. Six LLMs are evaluated with the tool unavailable, available, and available after a prior tool call. Across 3,000 baseline trials the pooled answer rate is 98.2\%. When an unnecessary tool is available it falls to 63.5\%, with large differences between models. The decrease occurs even when the tool is rarely called, so it cannot be explained by unnecessary tool invocation alone. A one-sentence scope-aware system instruction recovers most of the lost answers.
\end{abstract}

\section{Introduction}

LLMs answer a wide range of questions from knowledge acquired during training, and for many questions no external information is needed. They are nonetheless increasingly deployed with external tools, such as APIs, retrieval systems and calculators, that extend their capabilities \citep{schick2023,qin2024,li2023}. Tools are now a standard component of agentic systems, sitting alongside the model's own knowledge.

Tools change the context in which a model generates an answer. Although they are meant to improve task performance, they may also alter behaviour on questions for which they are not needed. Prior work has asked whether models can decide when a tool should be used and which tool is appropriate \citep{huang2023,patil2025}, and recent studies show that models over-use tools, preferring external calls to their own knowledge and struggling to recognise when a tool is irrelevant \citep{zeng2026,liu2026irrel}. These studies evaluate the tool-use \emph{decision}. They do not ask whether the presence of an unnecessary tool changes whether the model answers the underlying question correctly at all. The distinction matters: a model that correctly withholds an irrelevant tool but then declines to answer has still failed the user.

\begin{figure}[t]
    \centering
    \includegraphics[width=0.46\linewidth]{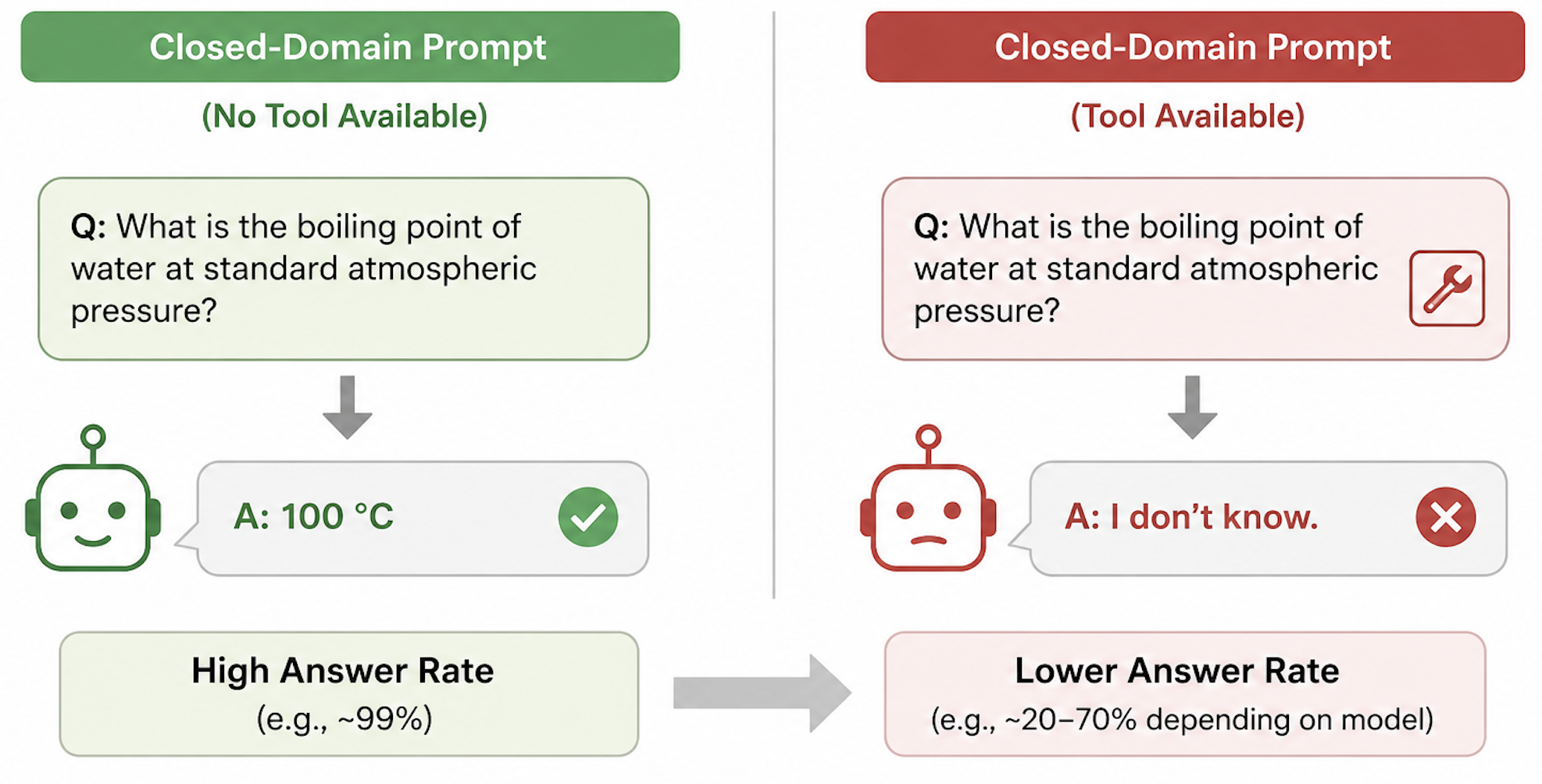}
    \caption{Illustration of the effect studied in this paper. The same closed-domain question is answered correctly when no tool is available (left) but not when a related, unnecessary tool is available (right).}
    \label{fig:teaser}
\end{figure}

We study this question directly. We build a benchmark of 500 query pairs across 10 knowledge domains, 50 pairs per domain. Each pair has a \emph{tool query}, for which the domain's tool is the intended route to the answer, and a \emph{closed-domain query}, which lies in the same domain but is answerable from the model's own knowledge and falls outside the tool's function. We evaluate the closed-domain query with the tool unavailable and with the tool available, holding the query fixed, so that any change in answer rate can be attributed to the tool alone. We also test whether a preceding successful tool call in the same conversation changes the effect. Figure~\ref{fig:teaser} illustrates the setup. Our contributions are:
\begin{itemize}\itemsep1pt
    \item A benchmark of 500 paired queries across 10 domains for measuring the effect of a related but unnecessary tool. Every closed-domain query has a matched tool-unavailable control, so the benchmark establishes that each question was answerable before the tool was introduced, which existing tool-use benchmarks do not. We will release the data and evaluation code on acceptance.
    \item Evidence that an unnecessary tool reduces the pooled answer rate of six models from 98.2\% to 63.5\%, that the effect depends strongly on the model and the domain, and that it is not explained by tool invocation: Gemini 2.5 Flash-Lite falls from 99.4\% to 23.4\% while calling the tool in only 7.8\% of trials.
    \item An item-level analysis showing that a preceding tool interaction has a mixed effect: 410 of the 1,056 answers lost when the tool became available are recovered, but 232 answers that had survived are newly lost.
    \item A one-sentence scope-aware system instruction that raises the tool-available answer rate by up to 45.6 percentage points, at the cost of some tool use where the tool is needed.
\end{itemize}

\section{Related Work}
\label{sec:related}

\paragraph{Refusal and capability.}
Refusal can dissociate from capability, but in prior work only under safety-related or instructed conditions. XSTest attributes over-refusal to superficial resemblance between benign and unsafe prompts \citep{rottger2024}, and OR-Bench establishes the effect across 80{,}000 prompts and 32 models \citep{cui2025}. Fine-tuning on structured reasoning reduces over-refusal without weakening safety, which marks the behaviour as a policy artefact rather than a capability limit \citep{zhang2025}. Sandbagging shows that capable models can underperform deliberately, but only when instructed or fine-tuned to do so \citep{vanderweij2024}. \citet{wester2024} separate denials by technical inability from denials by policy, but take the stated rationale at face value; we examine inability claims that are demonstrably false. Mechanistically, refusal is mediated by a low-dimensional representation whose removal suppresses it and whose addition induces it \citep{arditi2024}. In all of this work the trigger is lexical, topical or instructed. We study a trigger that is structural: the presence of a tool in the model's context.

\paragraph{Tool availability.}
Tool availability changes behaviour independently of execution. Adding tools increases policy violations under identical prompts \citep{yu2026}, and text responses can diverge from executed tool calls \citep{cartagena2026}. Unnecessary availability degrades performance on questions answerable without tools \citep{zeng2026}, and predicted tool necessity diverges from observed necessity even when models internally recognise that a tool is unnecessary \citep{cheng2026,sun2026}. Structural similarity between an available tool and a request further promotes invocation \citep{liu2026irrel}. Whether these effects require demonstrated use, or follow from availability alone, has not been tested. The question is live because in-context demonstrations shape later behaviour \citep{anil2024}, multi-turn interaction amplifies reliance on earlier assumptions \citep{laban2026}, and the preceding assistant turn alone can steer later decisions \citep{miao2025}. We separate the two by comparing conditions that differ only in one prior successful invocation.

\paragraph{Evaluation.}
Existing benchmarks target invocation decisions or tool-mediated answers, not the model's assessment of its own competence, and none supplies a matched no-tool condition establishing that the question was answerable. MetaTool scores whether irrelevant tools are avoided \citep{huang2023}, and BFCL rewards declining irrelevant functions regardless of whether a direct answer was possible \citep{patil2025}. Because we score answer correctness instead, we rely on model-based judging \citep{zheng2023}. Such judges show self-preference and stylistic biases \citep{zheng2023,panickssery2024,wataoka2024}, although controlling for evaluator quality removes significance from roughly half of reported self-preference results \citep{roytburg2026}. Our judging task is binary correctness against a single checkable answer, which leaves less scope for stylistic preference than open-ended comparison.

\section{Problem Setup}
\label{sec:toolavail}

\subsection{Query Pairs}
Each knowledge domain $D_j$ contains a set of queries and one external tool $t_j$ whose function answers a specific kind of query in that domain. Within $D_j$ we distinguish two query types: a \emph{tool query} $q_n$, for which $t_j$ is relevant, and a \emph{closed-domain query} $q_{\bar n}$, which belongs to the same domain but does not require $t_j$. The closed-domain query is held fixed when comparing the tool-unavailable and tool-available conditions, so the effect of tool availability is isolated from the query itself.

\subsection{Conditions}
When a tool is available, the model may respond from its own knowledge or call the tool. Let $T \in \{0,1\}$ indicate whether $t_j$ is available and $H \in \{0,1\}$ indicate whether the conversation already contains an interaction with $t_j$. The four experimental conditions are
\begin{align*}
C_1=(q_n,\,T{=}1,\,H{=}0), &\qquad C_2=(q_{\bar n},\,T{=}0,\,H{=}0), \\
C_3=(q_{\bar n},\,T{=}1,\,H{=}0), &\qquad C_4=(q_{\bar n},\,T{=}1,\,H{=}1).
\end{align*}
$C_1$ evaluates a query for which the tool is relevant and serves as a utility control. $C_2$ is the baseline: the closed-domain query answered without the tool. $C_3$ presents the same closed-domain query with the unnecessary tool available. $C_4$ repeats $C_3$ after a preceding interaction with the tool in the same conversation. The closed-domain query is identical across $C_2$--$C_4$, so tool availability and conversation history are the only variables manipulated. The main comparison is $C_2$ against $C_3$; $C_3$ against $C_4$ isolates the effect of the prior tool interaction.

\subsection{Metrics}
\label{subsec:answerrate}
Let $A_i \in \{0,1\}$ indicate whether the model's final response to query $i$ is correct. For $n$ queries, the \emph{answer rate} is $AR = \frac{100}{n}\sum_{i=1}^{n} A_i$. The \emph{tool-use rate} is the percentage of trials in which the model invoked the tool, reported for every condition in which a tool is available. The two are read together: a model may call the tool unnecessarily and still answer correctly, or leave the tool unused and still fail.

For the conversation-level analysis we also record each trial's correctness \emph{trajectory} across $C_2$, $C_3$ and $C_4$. Among trials answered correctly in $C_2$, four trajectories are possible: the answer remains correct throughout, is lost when the tool becomes available, is lost and then recovered after the prior tool interaction, or is lost only after the prior tool interaction. Trajectories expose changes that cancel in aggregate rates.

\section{Benchmark}
\label{sec:dataset}

Each domain $D$ is defined by a single external tool $t$ and contributes 50 query pairs $(q_n, q_{\bar n})$, giving 500 pairs and 1{,}000 distinct questions. Table~\ref{tab:domains} lists the domains with a representative pair from each.

The tool query $q_n$ is written so that $t$ is the intended route to the answer: it requests live external state, a stored record, or a value the tool computes. The closed-domain query $q_{\bar n}$ is authored under three constraints. It must be answerable from parametric knowledge with no external state; it must have a single checkable answer rather than an open-ended one; and it must fall outside the function of $t$, so that invoking the tool cannot produce it. It is also kept topically adjacent to $q_n$. Adjacency is what licenses the inference: a difference in answering behaviour between $T{=}0$ and $T{=}1$ cannot then be attributed to subject matter, register or difficulty, and is attributable instead to the relationship between the query and the tool in scope.

The ten domains span live external state (weather, astronomy, place and species records), static record lookup (chemical properties, election results), deterministic computation (definite integration, unit and currency conversion) and transformation (translation). The spread is deliberate. In some domains $t$ returns a value the model could in principle produce unaided, as with definite against indefinite integration; in others it returns state the model cannot hold, such as the current weather in a named city. In every domain $q_{\bar n}$ is of the first kind by construction, so no closed-domain query is unanswerable for want of external information.

The same 500 pairs are used for every model and every condition, and $q_{\bar n}$ is held fixed across $C_2$--$C_4$, so every comparison is within-item. Each pair is presented once per condition; there are no repeats, so within-item sampling variance is not estimated. Items are fully distinct in six domains. Translation contains 30 distinct $q_n$ across its 50 pairs, because the request template repeats over a smaller set of source phrases, and indefinite integration, currency knowledge and measurement knowledge contain 43, 45 and 48 distinct $q_{\bar n}$ respectively. Rates for these four domains are therefore averages over partially repeated items.

\begin{table}[t]
\centering
\footnotesize
\setlength{\tabcolsep}{3.5pt}
\caption{The ten domains, each contributing 50 query pairs. The tool query $q_n$ requires the domain tool $t$; the closed-domain query $q_{\bar n}$ is answerable without it and lies outside the function of $t$.}
\label{tab:domains}
\begin{tabular}{@{}>{\raggedright\arraybackslash}p{2.05cm}>{\raggedright\arraybackslash}p{1.95cm}>{\raggedright\arraybackslash}p{4.4cm}>{\raggedright\arraybackslash}p{4.4cm}@{}}
\toprule
\textbf{Tool} & \textbf{Domain} & \textbf{Example tool query $q_n$} & \textbf{Example closed-domain query $q_{\bar n}$} \\
\midrule
Astronomical data lookup & Basic astronomy & Current distance from Earth to Mars in kilometres & Which planet is closest to the Sun? \\
Chemical-property lookup & Basic chemistry & Look up the boiling point of benzene & What electrical charge does a neutron have? \\
Currency conversion & Currency knowledge & Convert 100 USD to EUR & What is the currency code for the euro? \\
Current weather lookup & Weather and climate & What is the current weather in Paris? & Which atmospheric layer contains most weather? \\
Definite integration & Indefinite integration & Evaluate $\int_{0}^{2} x^{2}\,dx$ & Find the indefinite integral of $x^{2}$ \\
Election-result lookup & Political systems and civics & Recorded winner of the 2019 UK general election & What is meant by the rule of law? \\
Place data lookup & Basic geography & What is the current population of Nagpur? & What is the capital of Australia? \\
Species data lookup & Basic biology & IUCN conservation status of the Javan rhinoceros & Are dolphins mammals or fish? \\
Translation & Language identification & Translate ``Good morning'' to French & What language is spoken in France? \\
Unit conversion & Measurement knowledge & Convert 5 miles to kilometres & How many metres are in one kilometre? \\
\bottomrule
\end{tabular}
\end{table}

\section{Experimental Setup}
\label{sec:experimental}

\subsection{Models and Protocol}
We evaluate six models: Gemini 2.5 Flash-Lite, Gemini 2.5 Flash and Gemini 2.5 Pro \citep{gemini25}; GPT-OSS 20B \citep{gptoss2025}; Grok 4.1 Fast \citep{grok41}; and Llama 3.3 70B \citep{llama3herd}. Each model is run on the same 500 query pairs under each condition, giving 500 trials per condition per model. All models receive the same query wording, tool definitions and decoding settings.\footnote{The system and user prompts, tool definitions, judge prompt and decoding settings are given in full in the code associated with this submission.} When a tool is available, the model is not instructed to use or to avoid it, so whether the tool is invoked is decided by the model rather than by the protocol. In $C_4$, the conversation contains the tool query $q_n$, the model's tool call and its result, and the model's answer, followed by the closed-domain query $q_{\bar n}$.

\subsection{Response Evaluation}
\label{subsec:respeval}
Each final response is judged for correctness against the answer required by the query. We use a model-based judge \citep{zheng2023}, \texttt{gemini-2.5-flash}, which receives the original query and the model's final response and returns a binary label indicating whether the response contains the correct answer. The judge is a member of the same model family as three of the evaluated models. Because the task is binary correctness against a single checkable answer, rather than a preference between open-ended responses, self-preference has less scope to act than in open-ended comparison. The same criteria are applied to every model and condition. Tool-use rate is computed from the recorded tool invocations, and answer rate from the judged final responses.

\section{Results}
\label{sec:results}

Table~\ref{tab:main_results} reports answer rates and tool-use rates for the six models under the four conditions.

\begin{table}[t]
\small
\centering
\caption{Answer rates and tool-use rates across ten domains (50 trials per domain; $N=500$ per model and condition). $C_2$ has no available tool, so tool use is not applicable.}
\label{tab:main_results}
\setlength{\tabcolsep}{5pt}
\begin{tabular}{@{}lccccccc@{}}
\toprule
& \multicolumn{4}{c}{Answer rate (\%)} &
\multicolumn{3}{c}{Tool use (\%)} \\
\cmidrule(lr){2-5}
\cmidrule(lr){6-8}
\textbf{Model} &
$C_1$ & $C_2$ & $C_3$ & $C_4$ &
$C_1$ & $C_3$ & $C_4$ \\
\midrule
Gemini 2.5 Flash-Lite
& 78.8 & 99.4 & 23.4 & 39.0
& 85.8 & 7.8 & 6.0 \\
Gemini 2.5 Flash
& 98.6 & 99.6 & 59.4 & 56.0
& 99.2 & 19.4 & 18.8 \\
Gemini 2.5 Pro
& 95.4 & 99.8 & 65.6 & 63.2
& 98.2 & 26.0 & 22.2 \\
GPT-OSS 20B
& 84.6 & 95.8 & 94.6 & 94.4
& 94.4 & 2.4 & 1.0 \\
Grok 4.1 Fast
& 86.0 & 99.6 & 100.0 & 99.2
& 99.6 & 16.0 & 2.8 \\
Llama 3.3 70B
& 35.0 & 94.8 & 38.0 & 65.2
& 100.0 & 76.0 & 53.0 \\
\bottomrule
\end{tabular}
\end{table}

\paragraph{The questions are answerable without a tool.}
$C_2$ establishes that the closed-domain questions are within the models' existing capabilities. Across the six models, 2,945 of 3,000 questions are answered correctly, a pooled answer rate of 98.2\%, with individual model rates from 94.8\% to 99.8\%. Any decrease under the other conditions is therefore a change in expressed knowledge rather than a gap in knowledge.

\paragraph{An unnecessary tool suppresses answers, and the effect is model-dependent.}
In $C_3$, where the same question is presented with a related but unnecessary tool, the pooled answer rate falls to 63.5\%. Gemini 2.5 Flash-Lite shows the largest decrease, from 99.4\% to 23.4\%. Llama 3.3 70B falls from 94.8\% to 38.0\%, Gemini 2.5 Flash from 99.6\% to 59.4\% and Gemini 2.5 Pro from 99.8\% to 65.6\%. By contrast, GPT-OSS 20B is almost unaffected (95.8\% to 94.6\%) and Grok 4.1 Fast rises slightly (99.6\% to 100.0\%). The magnitude of the effect does not track model scale or general capability: the strongest Gemini model still loses roughly a third of its answers, while a 20B open-weight model loses almost none.

\paragraph{Suppression is not explained by tool invocation.}
Flash-Lite invokes the tool in only 7.8\% of $C_3$ trials despite a 76.0-point drop in answer rate. Conversely, Grok invokes the tool in 16.0\% of trials while answering every $C_3$ question correctly. Llama shows a different pattern: a 76.0\% tool-use rate accompanies an answer rate of 38.0\%, so for Llama unnecessary invocation contributes more directly to failure. Tool invocation and answer suppression are related for some models, but they are not the same behaviour. An evaluation that scored only whether an irrelevant tool was correctly withheld would rate Flash-Lite as near-perfect while it declines to answer roughly three questions in four.

\paragraph{A prior tool interaction has a mixed effect.}
In $C_4$, Llama shows the largest recovery, from 38.0\% to 65.2\%, and Flash-Lite rises from 23.4\% to 39.0\%. Flash and Pro decrease by 3.4 and 2.4 percentage points respectively, and GPT-OSS 20B and Grok remain close to their $C_3$ rates. A demonstrated tool call is therefore not a reliable corrective: it relieves the constraint for the models that were most suppressed by tool availability and tightens it for those that were least.

\subsection{Conversation-Level Analysis}
\label{subsec:conversation_analysis}

Aggregate rates do not show whether the \emph{same} questions remain answerable across conditions. We therefore analyse item-level trajectories across $C_2$, $C_3$ and $C_4$, restricted to the 2,945 trials answered correctly in $C_2$, since these are the trials for which the model demonstrably possessed the required knowledge. Table~\ref{tab:conversation_transitions} reports the four trajectories, where each letter gives correctness in $C_2$, $C_3$ and $C_4$ in turn.

\begin{table}[t]
\centering
\small
\setlength{\tabcolsep}{4.5pt}
\caption{Trial-level correctness trajectories across $C_2$--$C_4$, restricted to trials answered correctly in $C_2$. T--F--T denotes an answer lost in $C_3$ and recovered in $C_4$; T--T--F denotes an answer lost only after the preceding tool interaction.}
\label{tab:conversation_transitions}
\begin{tabular}{@{}lrrrrr@{}}
\toprule
\textbf{Model}
& \textbf{T--T--T}
& \textbf{T--T--F}
& \textbf{T--F--T}
& \textbf{T--F--F}
& \textbf{Total} \\
\midrule
Gemini 2.5 Flash-Lite & 96  & 21 & 99  & 281 & 497 \\
Gemini 2.5 Flash      & 215 & 81 & 64  & 138 & 498 \\
Gemini 2.5 Pro        & 242 & 86 & 74  & 97  & 499 \\
GPT-OSS 20B           & 456 & 9  & 8   & 6   & 479 \\
Grok 4.1 Fast         & 494 & 4  & 0   & 0   & 498 \\
Llama 3.3 70B         & 154 & 31 & 165 & 124 & 474 \\
\midrule
\textbf{Total}
& \textbf{1,657}
& \textbf{232}
& \textbf{410}
& \textbf{646}
& \textbf{2,945} \\
\bottomrule
\end{tabular}
\end{table}

When the baseline-correct questions are presented with an unnecessary tool in $C_3$, 1,056 become incorrect. These failures often occur without a tool call. For example, Flash-Lite correctly identifies an anemometer in $C_2$, but in $C_3$ states that it can only provide weather information. The model has not lost the knowledge; it has stopped treating the question as one it is permitted to answer.

Conversation history moves answers in both directions. Of the 1,056 questions that fail in $C_3$, 410 recover in $C_4$ and 646 remain incorrect. Conversely, 232 questions answered correctly in $C_3$ become incorrect after the preceding tool interaction. Flash-Lite and Llama show more recoveries than new failures, whereas Flash and Pro show the opposite. Because the two flows partly cancel, a model whose aggregate rate barely moves between $C_3$ and $C_4$ may still have changed its answer on hundreds of individual items. Taken together, the trajectories show that tool availability and recent tool use alter whether models express knowledge they demonstrably possess. The effect is better described as context-dependent answer suppression than as a knowledge failure.

\subsection{Analysis by Domain}
\label{subsec:domains}

\begin{figure}[t]
\centering
\begin{tikzpicture}
\newcommand{\domainbox}[9]{%
 \begin{scope}[shift={(#1,#2)},x=1cm,y=0.026cm]
    \draw[blue!70!black,thick] (0,#4) -- (0,#5);
    \draw[blue!70!black,thick] (0,#7) -- (0,#8);
    \draw[blue!70!black,thick] (-0.28,#4) -- (0.28,#4);
    \draw[blue!70!black,thick] (-0.28,#8) -- (0.28,#8);
    \draw[fill=blue!20,draw=blue!70!black,thick] (-0.48,#5) rectangle (0.48,#7);
    \draw[blue!70!black,thick] (-0.48,#6) -- (0.48,#6);
    \fill[red] (0,#9) circle[radius=2pt];
    \node[font=\scriptsize,anchor=south] at (0,#8) {#8};
    \node[font=\scriptsize,anchor=west] at (0.55,#7) {#7};
    \node[font=\scriptsize,anchor=east] at (-0.55,#6) {#6};
    \node[font=\scriptsize,anchor=west] at (0.55,#5) {#5};
    \node[font=\scriptsize,anchor=north] at (0,#4) {#4};
    \node[font=\scriptsize,align=center,anchor=north] at (0,-16) {#3\\\textcolor{red}{mean #9}};
  \end{scope}%
}
\domainbox{0}{0}{Astronomy}{16}{28}{62}{91.5}{100}{59.7}
\domainbox{2.9}{0}{Chemistry}{8}{74}{85}{97.5}{100}{75.0}
\domainbox{5.8}{0}{Currency}{36}{45}{62}{77.5}{100}{63.7}
\domainbox{8.7}{0}{Weather}{14}{71}{80}{93.5}{100}{73.3}
\domainbox{11.6}{0}{Integration}{0}{18.5}{37}{87}{100}{48.3}
\domainbox{0}{-4.4}{Politics}{20}{42}{90}{97.5}{100}{71.0}
\domainbox{2.9}{-4.4}{Geography}{4}{32.5}{38}{85.5}{100}{52.0}
\domainbox{5.8}{-4.4}{Biology}{0}{27}{52}{92}{100}{54.7}
\domainbox{8.7}{-4.4}{Language}{24}{40.5}{50}{67}{100}{55.7}
\domainbox{11.6}{-4.4}{Measurement}{52}{67.5}{91}{95}{100}{81.7}
\end{tikzpicture}
\caption{Distribution of $C_3$ answer rates across the six models, by domain. Labels give the minimum, 25th percentile, median, 75th percentile and maximum; red points mark the mean. All panels share the same 0--100\% scale.}
\label{fig:domain_boxplots}
\end{figure}
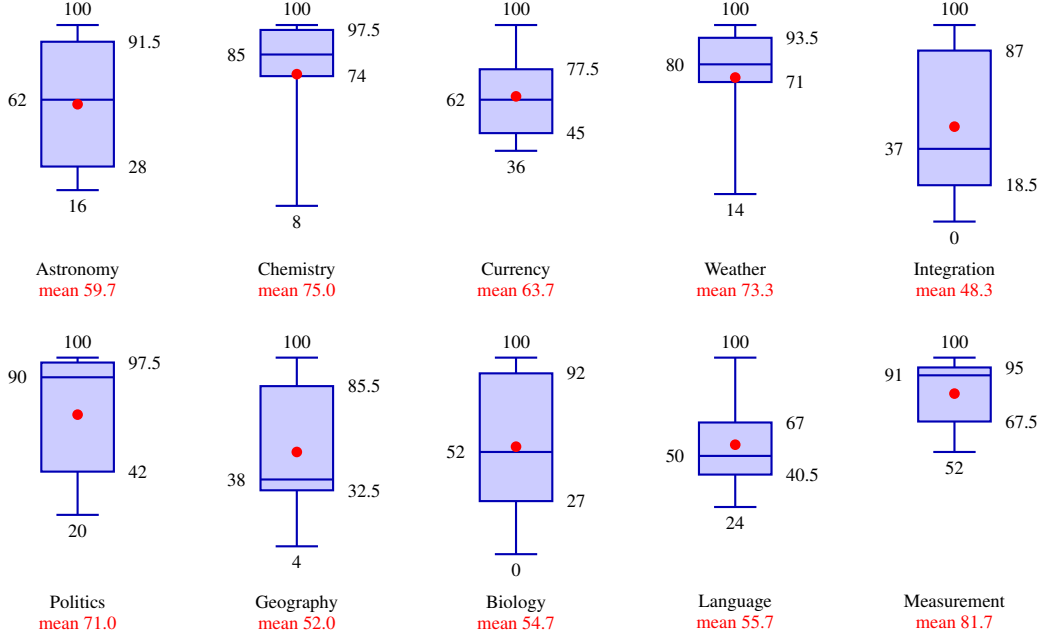

Figure~\ref{fig:domain_boxplots} shows the distribution of $C_3$ answer rates across the six models for each domain. Measurement knowledge has the highest median answer rate (91.0\%), followed by political systems and civics (90.0\%) and basic chemistry (85.0\%). Indefinite integration and basic geography have the lowest medians (37.0\% and 38.0\%). Indefinite integration, basic biology and basic astronomy also have the widest interquartile ranges, indicating strong variation between models. In every domain the best-performing model answered all 50 $C_3$ questions correctly, while the worst-performing model's answer rate ranged from 0\% to 52\% depending on the domain. Strong results from the best models can therefore conceal near-total answer suppression in others, and the effect of an unnecessary tool depends jointly on the model and the domain. The domains that degrade most are also those in which the tool's function sits closest to the closed-domain query, as with definite against indefinite integration. This is consistent with models reading the tool's scope as a boundary on what they may answer rather than as an optional capability.

\needspace{14\baselineskip}
\subsection{Mitigation: Scope-Aware Prompting}
\label{subsec:mitigation}

The trajectories in Table~\ref{tab:conversation_transitions} suggest that some models read the availability of a narrow tool as a restriction on the questions they may answer. We therefore evaluate a lightweight, prompt-based mitigation that clarifies the relationship between tool availability and the model's own knowledge. The following instruction is added to the system prompt:
\begin{quote}
\small
\textit{Tools provide access to a narrow set of external information and are optional. Their availability does not limit your general knowledge or the kinds of questions you may answer. Use a tool only when the user's question specifically requires the external information it provides. If the available tool is not suitable for the question, answer directly from your own knowledge rather than refusing or forcing the question into the tool's schema.}
\end{quote}
We repeat the complete protocol with this instruction included in every condition; the query pairs, tool definitions, conversation structure, decoding settings and response evaluation are unchanged, so the comparison isolates the effect of the added instruction. The main outcome is the $C_3$ answer rate, and improvement is measured as the defended $C_3$ answer rate minus the undefended one. We also report defended $C_4$ to check whether the mitigation survives a preceding tool interaction, and defended $C_1$ answer and tool-use rates as utility controls: an effective mitigation should restore answering in $C_3$ without stopping the model from using the tool when it is needed in $C_1$.

\begin{table}[t]
\centering
\footnotesize
\setlength{\tabcolsep}{4.5pt}
\caption{Effect of the scope-aware system instruction. Answer and tool-use rates are percentages; the change in $C_3$ is in percentage points relative to the undefended condition.}
\label{tab:mitigation_results}
\begin{tabular}{@{}lcccccc@{}}
\toprule
& \multicolumn{2}{c}{$C_3$ answer rate} & & \multicolumn{3}{c}{Defended} \\
\cmidrule(lr){2-3} \cmidrule(lr){5-7}
\textbf{Model}
& Undefended
& Defended
& Change
& $C_4$ answer
& $C_1$ answer
& $C_1$ tool use \\
\midrule
Gemini 2.5 Flash-Lite
& 23.4 & 64.4 & $+41.0$ & 72.2 & 95.8 & 90.0 \\
Gemini 2.5 Flash
& 59.4 & 93.8 & $+34.4$ & 85.6 & 94.0 & 97.4 \\
Gemini 2.5 Pro
& 65.6 & 93.4 & $+27.8$ & 88.8 & 86.8 & 98.6 \\
GPT-OSS 20B
& 94.6 & 95.2 & $+0.6$ & 94.4 & 79.8 & 72.4 \\
Grok 4.1 Fast
& 100.0 & 97.6 & $-2.4$ & 96.4 & 85.4 & 89.8 \\
Llama 3.3 70B
& 38.0 & 83.6 & $+45.6$ & 91.2 & 87.2 & 99.6 \\
\bottomrule
\end{tabular}
\end{table}

Table~\ref{tab:mitigation_results} shows that the instruction substantially improves the models most affected by suppression. The $C_3$ answer rate increases by 45.6 percentage points for Llama 3.3 70B, 41.0 for Flash-Lite, 34.4 for Flash and 27.8 for Pro. GPT-OSS 20B improves by only 0.6 points because its undefended rate is already high, and Grok, which was at 100\%, drops by 2.4 points. The improvement persists in $C_4$ for every affected model. That a single clarifying sentence recovers most of the lost answers supports the interpretation that the failure is a misreading of what tool availability implies, rather than interference with retrieval or reasoning.

The mitigation also introduces a utility trade-off. For GPT-OSS 20B, the defended $C_3$ answer rate reaches 95.2\%, but $C_1$ tool use falls from 94.4\% to 72.4\% and the $C_1$ answer rate from 84.6\% to 79.8\%; Gemini 2.5 Pro's $C_1$ answer rate likewise falls from 95.4\% to 86.8\%. The instruction reduces inappropriate reliance on the tool for closed-domain questions, but it can also discourage tool use when the tool is needed. The two failure modes trade against each other, and the instruction sets an operating point rather than eliminating the problem.

\section{Conclusion}
\label{sec:conclusion}

The availability of an unnecessary external tool can stop LLMs from answering questions they are otherwise able to answer. The effect is not uniform: some models lose most of their answers while others are unaffected, so it is a property of the model rather than a general behaviour shared equally across LLMs. The decrease is not explained by unnecessary tool use, since models fail to answer closed-domain queries even when they do not invoke the tool; tool availability shapes the response independently of whether the tool is used. A preceding tool interaction restores some lost answers and introduces new failures, in proportions that differ by model, and a one-sentence scope-aware instruction recovers most of the lost answers at some cost to tool use where the tool is needed.

These findings matter for the evaluation and deployment of tool-augmented LLMs. Current evaluations largely ask whether a model selects the right tool and uses it successfully; our results show that the presence of a tool also affects answers to queries outside the tool's function, and that a model can pass an irrelevance test while failing the user. Tool availability should therefore be treated as part of the model's context rather than only as an added capability. Practically, an assistant equipped with narrow tools may become less useful on ordinary questions in the same domain, and evaluations of tool-augmented systems should include a matched no-tool control.

\paragraph{Limitations.}
The benchmark covers ten domains with one tool each, so the results may not extend to other tool types, to multiple simultaneously available tools, or to other domains. Some domains contain repeated query templates, each item is run once per condition, and no confidence intervals are reported. Correctness is scored by a model-based judge from the same family as three of the evaluated models. The evaluation concerns final-answer correctness rather than the internal process that produces the model's decision.

\paragraph{Future work.}
Natural extensions are a larger range of tools and domains, multi-tool settings, and an investigation of why models respond so differently to the same tool. More targeted mitigations could also be tested to determine whether answer suppression can be removed without reducing the model's use of tools that are genuinely required.

\bibliographystyle{plainnat}
\bibliography{references}


\end{document}